\documentclass[10pt,journal,compsoc]{IEEEtran}

\ifCLASSOPTIONcompsoc
  \usepackage[nocompress]{cite}
\else
  \usepackage{cite}
\fi
\ifCLASSINFOpdf
\else
\fi
\usepackage{amsmath}
\usepackage{algorithmic}
\usepackage{array}
\usepackage{url}
\usepackage{times}
\usepackage{epsfig}
\usepackage{graphicx}
\usepackage{amsmath}
\usepackage{bm}
\usepackage{amssymb}
\usepackage{breqn}
\usepackage{subcaption}
\usepackage{verbatim}
\usepackage{algorithm2e}
\usepackage{multirow}
\usepackage{array}
\usepackage{color}
\usepackage{wrapfig}
\usepackage[%
    colorlinks=true,
    pdfborder={0 0 0},
    linkcolor=red
]{hyperref}

\newcommand{\Blue}[1]{\textcolor{black}{#1}}

\newcommand{\boldstart}[1]{\noindent\textbf{#1}}
\newcolumntype{P}[1]{>{\centering\arraybackslash}p{#1}}
\newcommand{\argmin}{\arg\!\min}

\begin{document}
\title{Spatiotemporal Bundle Adjustment for Dynamic 3D Human Reconstruction in the Wild}

\author{Minh~Vo, Yaser~Sheikh, and Srinivasa~G.~Narasimhan
\IEEEcompsocitemizethanks{\IEEEcompsocthanksitem The authors are with The Robotics Institute, Carnegie Mellon University, PA, USA, 15213.\protect\\
E-mail:\{mpvo, srinivas, yaser\}@cs.cmu.edu}}
\markboth{TRANSACTION ON PATTERN ANALYSIS AND MACHINE INTELLIGENCE, VOL. XXX, NO. XXX, XXX XXX}%
{Minh~Vo \MakeLowercase{\textit{et al.}}: Bare Demo of IEEEtran.cls for Computer Society Journals}

\IEEEtitleabstractindextext{%
\begin{abstract}
Bundle adjustment jointly optimizes camera intrinsics and extrinsics and 3D point triangulation to reconstruct a static scene. The triangulation constraint, however, is invalid for moving points captured in multiple unsynchronized videos and bundle adjustment is not designed to estimate the temporal alignment between cameras. We present a spatiotemporal bundle adjustment framework that jointly optimizes four coupled sub-problems: estimating camera intrinsics and extrinsics, triangulating static 3D points, as well as sub-frame temporal alignment between cameras and computing 3D trajectories of dynamic points. Key to our joint optimization is the careful integration of physics-based motion priors within the reconstruction pipeline, validated on a large motion capture corpus of human subjects. We devise an incremental reconstruction and alignment algorithm to strictly enforce the motion prior during the spatiotemporal bundle adjustment. This algorithm is further made more efficient by a divide and conquer scheme while still maintaining high accuracy. We apply this algorithm to reconstruct 3D motion trajectories of human bodies in dynamic events captured by multiple uncalibrated and unsynchronized video cameras in the wild. To make the reconstruction visually more interpretable, we fit a statistical 3D human body model to the asynchronous video streams. 
Compared to the baseline, the fitting significantly benefits from the proposed spatiotemporal bundle adjustment procedure. Because the videos are aligned with sub-frame precision, we reconstruct 3D motion at much higher temporal resolution than the input videos.
\textbf{Website}: \url{http://www.cs.cmu.edu/~ILIM/projects/IM/STBA}
\end{abstract}

}

\maketitle

\IEEEdisplaynontitleabstractindextext
\IEEEpeerreviewmaketitle

\IEEEraisesectionheading{\section{Introduction}\label{sec:introduction}}

When a moving point is observed from multiple cameras with simultaneously triggered shutters, the dynamic 3D reconstruction problem reduces exactly to the case of static 3D reconstruction. The classic point triangulation constraint~\cite{longuet1981computer}, and the algorithmic edifice of bundle adjustment~\cite{triggs2000bundle} built upon it, applies directly. Currently, there exists no consumer mechanism to ensure that multiple cameras, i.e., smartphones, consumer camcorders, or egocentric cameras, are simultaneously triggered~\cite{latimer2014socialsync}. Thus, in the vast majority of dynamic scenes captured by multiple independent video cameras, no two cameras see the 3D point at the same time instant. This fact trivially invalidates the triangulation constraint.

To optimally solve the dynamic 3D reconstruction problem, we must first recognize all the constituent sub-problems that exist. The classic problems of point triangulation and camera resectioning in the static case are subsumed. In addition, two new problems arise: reconstructing 3D trajectories of moving points and estimating the temporal location of each camera. Second, we must recognize that the sub-problems are tightly coupled. As an example, consider the problem of estimating 3D camera pose. Prior work segments out the stationary points and uses them to estimate the camera pose~\cite{park2015Trajectory}. However, this approach ignores evidence from moving points that are often closer to the cameras and therefore provide tighter constraints for precise camera calibration.
Imprecise camera calibration and quantization errors in estimating discrete temporal offsets result in significant errors in the reconstruction of moving points\footnote{Consider a person jogging at 10m/s. He is captured by two cameras at 30Hz, one static and one handheld jittering at 3mm per frame, with the camera baseline of 1m, and from 4m away. A simple calculation suggests that a n{\"a}ive attempt to triangulate points of the static camera with their correspondences of the best-aligned frame in the other camera results in up to 40 cm reconstruction error.}~\cite{raguse2006photogrammetric,furukawa2010accurate,vo2012hyper}.

Prior work in dynamic 3D reconstruction has addressed some subset of these problems. For instance, assuming known (or separately estimated) camera pose and temporal alignment, Avidan and Shashua pose the problem of trajectory triangulation~\cite{avidan2000trajectory}, where multiple noncoincidental projections of a point are reconstructed. Trajectory triangulation is an ill-posed problem and current algorithms appeal to motion priors to constrain reconstruction: linear and conical motion~\cite{avidan2000trajectory}; smooth motion~\cite{park2015Trajectory,valmadre2012general}; sparsity priors ~\cite{Zhu2015sparsetrajectory}; low rank spatiotemporal priors ~\cite{simon2014separable}. Estimating the relative temporal offsets of videos captured by the moving cameras is more involved ~\cite{yan2004video,gaspar2014synchronization}. Currently, the most stable temporal alignment methods require corresponding 2D trajectories as input ~\cite{wedge2006motion,caspi2006feature, padua2010linear, elhayek2012feature, tresadern2009video} and rely purely on geometric cues to align the interpolated points along the trajectories across cameras. Recent work has considered the aggregate problem but addresses the spatial and temporal aspects of the problem independently~\cite{basha2012photo, ji2016spatio, zheng2017self}. 

In this paper, we introduce the novel concept of spatiotemporal bundle adjustment that jointly optimizes all sub-problems simultaneously. Just as with static 3D reconstruction, where the most accurate results are obtained by jointly optimizing for camera parameters and triangulating static points, the most accurate results for dynamic 3D reconstruction are obtained when jointly optimizing for the spatiotemporal camera parameters and triangulating both static and dynamic 3D points. Unlike traditional bundle adjustment, we recognize the need for a motion prior in addition to the standard reprojection cost to jointly estimate the 3D trajectories corresponding to the sub-frame camera temporal alignment. Such joint estimation is most helpful for dynamic scenes with large background/foreground separation where the spatial calibration parameters estimated using background static points are unavoidably less accurate for foreground points. We evaluate several physics-based 3D motion priors (least kinetic energy, least force, and least action) on the CMU motion capture repository~\cite{CMU-Mocap} for spatiotemporal bundle adjustment.

Direct optimization of the spatiotemporal objective is hard and is susceptible to local minima. We address this optimization problem using an incremental reconstruction and temporal alignment algorithm. This optimization framework ensures the proposed 3D motion prior constraint is satisfied. Our algorithm naturally handles the case of missing data (e.g., when a point is occluded in a particular time instant). This algorithm enables accurate 3D trajectory estimation at much higher temporal resolution than the frame rates of the input videos. 

The incremental reconstruction and alignment approach is effective and accurately optimizes the spatiotemporal bundle adjustment problem. However, its computational complexity grows quadratically with the number of cameras. We solve this issue by dividing the optimization problem into overlapping groups of cameras with overlapping field of view, each of which is optimized independently using the incremental reconstruction and alignment scheme. These sub-problems are merged and globally optimized in the final pass. Empirically, this approach is at least 10 times faster while suffers marginal accuracy loss on our datasets.

\Blue{Naturally, the spatiotemporal bundle adjustment algorithm relies on accurate timing of when the object is observed in each frame. However, for the commonly-used rolling shutter camera, each image row/column is exposed at slightly different time and has different pose. These factors must be accounted for to ensure accurate spatiotemporal calibration. While spatial pose estimation for rolling shutter camera are relatively well-studied~\cite{alblr6p,ovren2018spline,ito2017self,oth2013rolling}, calibrating the rolling shutter scanning rate is less explored and mostly requires specific calibration tool~\cite{geyer2005geometric}. To this end, we introduce a self-calibration method to estimate the rolling shutter readout speed directly from the observed scene without additional hardware. Our formulation shares the same principle as motion prior formulation: constant velocity within a short time (one frame). This approach noticeably improves the spatiotemporal bundle adjustment framework on real sequences.} 

We apply spatiotemporal bundle adjustment to reconstruct human dynamic scenes captured in the wild with rolling shutter smartphone cameras. While this algorithm can accurately reconstruct the 3D trajectory of the dynamic points, those points are usually sparse and visually hard to interpret. Thus, we fit a statistical 3D human body model~\cite{loper2015smpl} to the unsynchronized videos to augment the visualization. This is in a similar spirit of multiview stereo to sparse bundle adjustment~\cite{schonberger2016pixelwise, furukawa2010towards}. \Blue{We note the existence of prior work with triangulation-based synchronized multiview setups~\cite{joo2018total,huang2017towards}. This paradigm computes distorted body shape and pose as it wrongly aggregates information from unsynchronized cameras. Other approaches using single-frame~\cite{bogo2016keep,lassner2017unite,kanazawa2018end} or monocular video~\cite{huang2017towards,tung2017self} suffer from depth ambiguity which makes it difficult to merge the results across different cameras. Our solution is to use spatiotemporal calibration for accurate frame sequencing\footnote{\Blue{Sequencing refers to the temporal ordering of a set of frames~\cite{basha2012photo}}} and motion priors to link single-frame estimations to reconstruct the human body over the entire multiview sequences. We highlight the importance of accurate frame sequencing in the supplementary video where the incorrect sequencing produces noticeable jitters and loops in to body motion.} Ideally we should re-optimize camera calibration parameters and 3D points jointly with the body shape and pose coefficients. However, because the extracted semantic cues are often imprecise, such re-optimization hurts the spatiotemporal bundle adjustment. Thus, we fix the estimated spatiotemporal parameters during the shape fitting. As a demonstration, we reconstruct 3D trajectories and human body shape of dynamic events captured outdoor by at least eight smartphones without any constraints.

This paper extends our previous work~\cite{vo2016spatiotemporal} in multiple aspects. First, we provide a more thorough evaluation of the motion priors. Second, we devise a divide and conquer algorithm to speed up the incremental reconstruction and alignment framework. Third, we introduce a self-calibration method to estimate the rolling shutter readout speed directly from the scene without additional hardware. These contributions enable us to build a framework for accurate human shape fitting from multiple unsynchronized and uncalibrated low frame rate video cameras.

\section{Motion Prior for Dynamic 3D Capture}

Consider the scenario of $C$ video cameras observing $N$ 3D points over time. The relation between the 3D point $X^n(t)$ and its 2D projection $x^n_c(f)$ on camera $c$ at frame $f$ is given by:
\begin{equation}
\label{eq:Proj}
\begin{bmatrix}x^n_c(f) \\1 \end{bmatrix} \equiv 
K_c(f) \begin{bmatrix}	R_c(f) &T_c(f)\\	\end{bmatrix}
\begin{bmatrix}	X^n(t) \\	1 \end{bmatrix}, 
\end{equation}
where $K_c(f)$ is the intrinsic camera matrix, $R_c(f)$ and $T_c(f)$ are the relative camera rotation and translation, respectively. For simplicity, we denote this transformation as $x^n_c(f) = \pi_c(f,X^n(t))$. The time corresponding to row $r_c$ at frame $f$ is related to the continuous global time $t$ linearly: $f = \alpha_c t + \beta_c + \gamma_c r_c$, where $\alpha_c$ and $\beta_c$ are the camera frame rate and time offset, $\gamma_c$ is the rolling shutter pixel readout speed. For global shutter camera, $\gamma_c$ is zero.\\

\boldstart{Image reprojection cost:} At any time instance, the reconstruction of a 3D point must satisfy Eq.~\ref{eq:Proj}. This gives the standard reprojection error $S_I$, which we accumulate over all 2D points observed by all $\textit{C}$ cameras for all frames $\textit{F}_c$:
\begin{equation}
\label{eq:ImgCost}
S_I = \sum_{c = 1}^{\textit{C}}\sum_{n = 1}^{\textit{N}} \sum_{f=1}^{\textit{F}_c}I_c^n(f) \bigg( \frac{\Vert  \pi_c(f,X^n(t))- x_c^n(f) \rVert}{\sigma_c^n(f)}\bigg)^2,
\end{equation}
where, $I_c^n(f)$ is a binary indicator of the point-camera visibility, and $\sigma_c^n(f)$ is a scalar, capturing the uncertainty in localizing $x_c^n(f)$ to $S_I$. Since the localization uncertainty of an image point $x_c^n(f)$ is proportional to its scale~\cite{zeisl2009estimation}, we use the inverse of the feature scale as the weighting term for each residual term in $S_I$.

However, Eq.~\ref{eq:ImgCost} is purely spatially defined and does not encode any temporal information about the dynamic scene. Any trajectory of a moving 3D point must pass through all the rays corresponding to the projection of that point in all views. Clearly, there are infinitely many such trajectories and each of these paths corresponds to a different temporal sequencing of the rays. Yet, the true trajectory must also correctly align all the cameras. This motivates us to investigate a motion prior that ideally estimates a trajectory that corresponds to the correct temporal alignment. The cost of violating such a prior $S_M$ can be then added to the image reprojection cost to obtain a spatiotemporal cost function that jointly estimates both the spatiotemporal camera calibration parameters and the 3D trajectories:
\begin{equation} 
\label{eq:STCost}
S = \argmin_{\mathbf{X}(t), \{\mathbf{K}, \mathbf{R}, \mathbf{t}\}, \boldsymbol\alpha,\boldsymbol\beta}  S_I +  S_M.
\end{equation}

Given multiple corresponding 2D trajectories of both the static and the dynamic 3D points $\{\mathbf{x}_c(t)\}$ for $C$ cameras, we describe how to jointly optimize Eq.~\ref{eq:STCost} for the 3D locations $\mathbf{X}(t)$, the spatial camera parameters at each time instant $\{K_c(f),R_c(f),T_c(f) \}$ and the temporal alignment between cameras $\boldsymbol \beta$. We assume the frame rate $\boldsymbol \alpha$ is known.


\subsection{Physics-based Motion Priors}
In this section, we investigate several forms of motion prior needed to compute $S_M$ in Eq.~\ref{eq:STCost}. We validate each of these priors on the entire CMU Motion Capture Database~\cite{CMU-Mocap} for their effectiveness on modeling human motion.

When an action is performed, its trajectories must follow the paths that minimize a physical cost function. This inspires the investigation of the following three types of priors: least kinetic energy, least force\footnote{We actually use the square of the resulting forces.}, and least action~\cite{feynman1963feynman}. See Fig.~\ref{fig:Mocap3D_T} for the formal definition of these priors. In each of these priors, $m$ denotes the mass of the 3D point, $g$ is the gravitational acceleration force acting on the point at height $h(t)$, and $v(t)$ and $a(t)$ are the instantaneous velocity and acceleration at time $t$, respectively. \Blue{According to the Cauchy$-$Schwarz inequality, the least kinetic energy prior encourages constant velocity motion, the least force prior promotes constant acceleration motion\footnote{The trivial zero solution is discarded due to the image reprojection cost.}. According to the Newton law of physics, the least action prior favors projectile motion.} While none of these priors hold for an active system where forces are arbitrarily applied during its course of action, we conjecture that the cumulative forces applied by both mechanical and biological systems are sparse and over a small duration of time, the true trajectory can be approximated by the path that minimizes the costs defined by our motion priors. Any local errors in the 3D trajectory, either by inaccurate estimation of points along the trajectory or wrong temporal sequencing between points observed across different cameras, produce higher motion cost. \\

\boldstart{Least kinetic motion prior cost:} We accumulate the cost over all $N$ 3D trajectories for all time instances $T^n$:
\begin{dmath}
\label{eq:KineticCostSync}
S_M = \sum_{n = 1}^{N}\sum_{i = 1}^{T^n-1}  w_n(t) \frac{m_n}{2}v_n(t^{i})^2(t^{i+1} - t^i),
\end{dmath}
where $\gamma_n(t)$ is the weighting scalar and $m_n$ is the point mass, assumed to be identical for all 3D points and set to be 1. We approximate the instantaneous speed $v(t^i)$ at time $t^i$ along the sequence $X^n(t)$ by a forward difference scheme, $v_n(t^i) \approx \Vert \frac{X^n(t^{i+1}) - X^n(t^i)}{t^{i+1} - t^i} \rVert$. We add a small constant $\epsilon$ to the denominator to avoid instability caused by 3D points observed at approximately the same time. Eq.~\ref{eq:KineticCostSync} is rewritten as:

\begin{dmath}
\label{eq:KineticCostSync2}
S_M = \sum_{n = 1}^{N}\sum_{i = 0}^{T^n-1} \frac{w_n(t)}{2}\Big\Vert\frac{X^n(t^{i+1})-X^n(t^i)}{t^{i+1} - t^i+ \epsilon}\Big\Vert^2(t^{i+1} - t^i),
\end{dmath}

Using the uncertainty $\sigma^n_c(f)$ of the 2D projection of 3D point $X_n(t)$, the weighting $w_n(t)$ can be approximated by a scaling factor that depends on the point depth $\lambda$ and the scale $\mu$, relating the focal length to the physical pixel size, as $w_n = \frac{\mu \lambda}{\sigma^n_c}$.  The least force and least action prior costs can be computed similarly.

\subsection{Evaluation on Motion Capture Data}
\begin{figure*}[t]
\includegraphics[width=\linewidth]{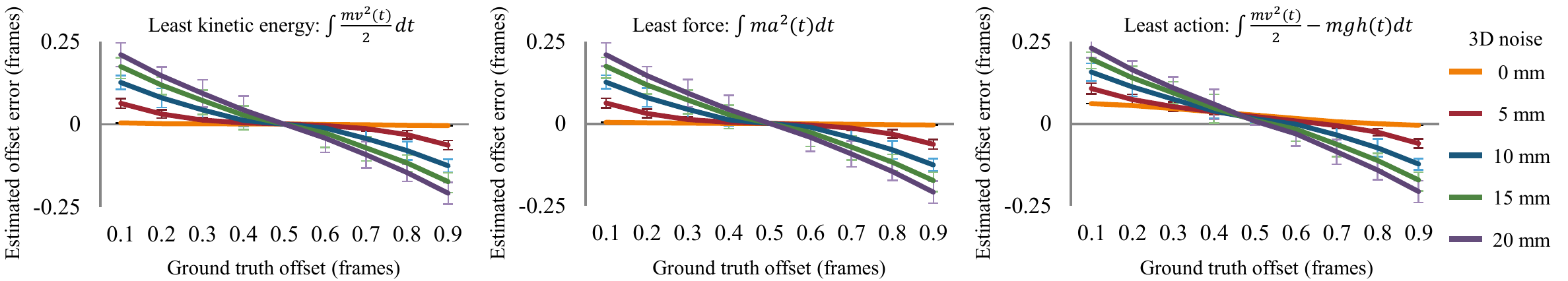}
\caption{Evaluation of the motion priors on 3D motion capture data. The least kinetic energy prior and least force prior performs similarly and both estimate the time offset between two noisy sequences obtained by uniformly sampling a 3D trajectory from different starting times. The least action prior gives biased results even for the no-noise case. }
\label{fig:Mocap3D_T}
\end{figure*}

Consider a continuous trajectory of a moving point in 3D.  Sampling this continuous trajectory starting at two different times produces two discrete sequences in 3D.
We first evaluate how the motion prior helps in estimating the temporal offset between the two discrete sequences. We extend this to 2D trajectories recorded by cameras later. The evaluation is conducted on the entire CMU Motion Capture Database~\cite{CMU-Mocap}, containing over 2500 sequences of common human activities such as playing, sitting, dancing, running and jumping, and captured at 120fps. 

Each trajectory is subsampled starting at two different random times to produce the discrete sequences. 3D zero-mean Gaussian noise is added to every point along the discrete trajectories. The ground truth time offsets are then estimated by a linear search and we record the solution with the smallest motion prior cost. For our test, the captured 3D trajectories are sampled at 12fps and the offsets are varied from  $0.1$ to $0.9$ frame interval in $0.1$ increments.

As shown in Fig.~\ref{fig:Mocap3D_T}, the least kinetic energy prior and least force prior perform similarly in this setting and both estimate the time offset between the two trajectories well for low noise levels. 
When more noise is added to the trajectory sequences, the sequencing is noisier. Yet, our motion cost favors correct camera sequencing over closer time offset. This is a desirable property because wrong sequencing results in a trajectory with loops (see Fig.~\ref{fig:Checker_Subframe}).
In contrast, the least action prior gives biased results even when no noise is added to the 3D data. 

\begin{wrapfigure}{L}{0.2\textwidth}
\includegraphics[width=0.2\textwidth]{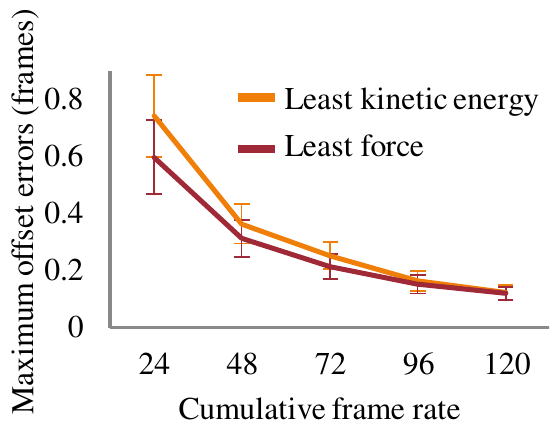}
\label{fig:motionvsframe}
\caption{Comparison of the sequencing expressiveness between the least kinetic energy prior and the least force prior for different cumulative frame rates.}
\end{wrapfigure} 

We further compare the sequencing expressiveness of the least kinetic energy prior to the least force prior for different cumulative frame rate. \Blue{The cumulative frame rate is defined as the frame rate of the virtual camera consisting of all the frames from each camera and captures the effective sampling rate of a multi-camera system.} As shown in Fig.~2, the least force prior is more expressive than the least kinetic prior for lower frame rate. This is expected since the least force prior captures more local information of the trajectory. However, for modern video cameras, the cumulative frame rate easily exceeds 120fps \Blue{(e.g, five 30fps cameras have the cumulative frame rate of 150fps)}, where the alignment results using either priors are similar. Thus, we only use the least kinetic prior in the remainder of the paper. Extension to the least force prior is straightforward.

\section{Spatiotemporal Bundle Adjustment} 

\begin{algorithm}[t]
\KwIn{$\{\mathbf{x}_c(t)\}, \{\mathbf{K'}, \mathbf{R'}, \mathbf{T'}\}, \boldsymbol\beta'$ }
\KwOut{$\{\mathbf{X}(t)_p\}, \{\mathbf{K}, \mathbf{R}, \mathbf{T}\}, \boldsymbol\beta$}
\textbf{1. (Sec. 3.1.1)} Refine the alignment pairwise \\
\textbf{2. (Sec. 3.1.2)} Generate prioritized camera list \\
\textbf{3. (Sec. 3.1.3)} \While{All cameras haved \textbf{NOT} been processed}
{
\For{All cameras slots}{
Solve Eq.~\ref{eq:STCost} for $\{\mathbf{X}_p(t)\}$ and $\boldsymbol\beta$\\
\eIf{No sequencing flipped}{
Record the STBA cost and its solution.
}{
Discard the solution;
}
}
Accept the solution with the smallest cost
}	
\textbf{4. (Sec. 3.2)} Solve Eq.~\ref{eq:STCost} for $\{\mathbf{X}(t)_p\}, \{\mathbf{K}, \mathbf{R}, \mathbf{T}\}, \boldsymbol\beta$ \\
\vspace{0.1in}
\caption{Incremental reconstruction and alignment.}\label{algo:inc}
\end{algorithm}

Unlike traditional bundle adjustment~\cite{triggs2000bundle}, the spatiotemporal bundle adjustment must jointly optimize for four coupled problems: camera intrinsics and extrinsics, 3D locations of static points, temporal alignment of cameras and 3D trajectories of dynamic points. However, direct optimization of Eq.~\ref{eq:STCost} is hard because: 
(a) it requires a solution to a combinatorial problem of correctly sequencing all the cameras and (b) motion prior cost is strongly discontinuous as small changes in time offsets can switch the temporal ordering of cameras. Thus, it is not possible to ensure the satisfaction of the motion prior constraint. 

We solve this problem using an incremental reconstruction and alignment approach where the camera is sequentially added to the optimization problem. This algorithm is further sped up by a divide and conquer scheme where the groups of cameras are solved independently first and then merged and refined globally using continuous second order optimization. We initialize temporal alignment and the 3D trajectory of the dynamic points using a geometry (or triangulation constraint) based method~\cite{snavely2006photo,elhayek2012feature}. Even though the triangulation constraint is not strictly satisfied, empirically, the estimations provide a good starting point for the incremental reconstruction and alignment. 

\subsection{Incremental Reconstruction and Alignment)}
Our incremental reconstruction and alignment (IRA) approach adds camera one at a time. For every new camera, a linear search for the best sequencing of this camera with respect to the previous cameras based on the motion prior cost is conducted. Once the sequencing order is determined, we use continuous optimization to jointly estimate all the spatiotemporal camera parameters, and static points and dynamic trajectories. Thanks to the linear search step, we can enforce the motion prior constraint strictly without any discontinuities due to incorrect time ordering of cameras. We summarize this method in Algorithm~\ref{algo:inc}. 

\subsubsection{Temporal alignment of two cameras} 
We refine the initial guess by optimizing Eq.~\ref{eq:STCost}. However, just as in point triangulation, the 3D estimation from a stereo pair is unreliable. Thus, we simply do a linear search on a discretized set of temporal offsets and only solve Eq.\ref{eq:STCost} for the 3D trajectories. The offset with the smallest cost is taken as the sub-frame alignment result. We apply this refinement to all pairs of cameras.

\subsubsection{Which camera to add next?}
As in incremental SfM~\cite{snavely2006photo,frahm2010building}, we need to determine the next camera to include in the calibration and reconstruction process.
For this, we create a graph with each camera as a node and define the weighted edge cost between any two cameras $i^{th}$ and $j^{th}$ as
\begin{equation} 
\label{eq:PrioritizedList}
E_{ij} = \sum_{k=1, k\neq i, j}^{C} S_{ij}{\frac{\vert t_{ij}+t_{jk}-t_{ik}\vert}{N_{ij}B_{ij}}},
\end{equation}
where $t_{ij}$, $N_{ij}$, $B_{ij}$, and $S_{ij}$ are the pairwise offset, the number of visible corresponding 3D points, the average camera baseline, and the spatiotemporal cost evaluated for those cameras, respectively. Intuitively, $\vert t_{ij}+t_{jk}-t_{ik}\vert$  encodes the constraint between the time offsets among a camera triplet, and $N_{ij}B_{ij}$ is a weighting factor favoring the camera pair with more common points and larger baseline. 

Similar to~\cite{elhayek2012feature,wang2014videosnapping}, a minimum spanning tree (MST) of the graph is used to find the alignment of all cameras. We use the Kruskal MST, which adds nodes with increasing cost at each step. The camera processing order is determined once from the connection step of the MST procedure. 

\subsubsection{Estimating the time offset of the next camera}
We temporally order the currently processed cameras and insert the new camera into possible time slots between them, followed by a nonlinear optimization to jointly estimate all the offsets and 3D trajectories. Any trials where the relative ordering between cameras change after the optimization are discarded, ensuring that the motion prior is satisfied. The trial with the smallest cost is taken as the temporal alignment and 3D trajectories of the new set of cameras. 

\subsection{Divide and Conquer}
While the incremental reconstruction and alignment approach offers a tractable solution for the loss function defined in Eq~.\ref{eq:STCost}, its complexity increases quadratically with the number of cameras. For every new camera, we must optimize Eq.~\ref{eq:STCost} $C-1$ times, where $C$ is the number of cameras being processed, to determine the sequencing order with the least motion cost. To address the computational efficiency issue, we propose a divide and conquer approach to speed up to solver while having minimal reconstruction accuracy loss. This algorithm is based on the observation that the temporal alignment becomes stable after a small number of cameras is processed (4 cameras in all of our experiments).

This algorithm proceeds in three stages. First, we form the camera groups with large co-visiblity with them by creating a skeleton graph using the camera graph built in Sec. 3.1.2~\cite{snavely2008skeletal}. Here, each group is taken as two camera nodes in the skeleton graph and their connected cameras in the graph of Sec. 3.1.2. We purposely let the overlapping groups share two cameras to better detect and discard temporal inconsistency when merging all the groups together. Second, we process each camera group independently processed using the incremental reconstruction and alignment approach. For every pair of inconsistency groups detected, these groups are merged and re-processed using the first approach. Third, we aggregate the temporal alignment parameters from all groups into a common timeline and optimize all cameras jointly for the spatiotemporal calibration parameters and the 3D position of the static and dynamic points. 

\subsection{Motion Resampling via Discrete Cosine Transform}
Note that Eq.~\ref{eq:KineticCostSync2} approximates the speed of the 3D point using finite difference. While this approximation allows better handling of missing data, the resulting 3D trajectories are often noisy. Thus, we further fit the weighted complete DCT basis function to the estimated trajectories. Our use of DCT for resampling is mathematically equivalent to our discrete motion prior~\cite{strang1999discrete} and is not an extra smoothing prior. For the uniform DCT resampling, the least kinetic energy prior cost \Blue{defined in  Eq.~\ref{eq:KineticCostSync}} is rewritten as:
\begin{dmath}
\label{eq:Reampling}
S_M = \sum_{n = 1}^{N}E^{n\top} W^n E^n \Delta t,
\end{dmath}
where $E^n$ is the DCT coefficient of the 3D trajectory $n$, $W^n$ is a predefined diagonal matrix, weighting the contribution of the bases, and $\Delta t$ is the resampling period. The 3D trajectory $X^n(t)$ is related to $E^n$ by  $X^n(t) = B^{n\top} E^n$, where $B^n$ is a predefined DCT basis matrix. The dimension of $B^n$ and $W^n$ depend on the trajectory length. We replace the trajectory $X^n(t)$ in Eq.~\ref{eq:ImgCost} by $B^{n\top} E^n$ and rewrite Eq.~\ref{eq:STCost} as:
\begin{dmath}
\label{eq:Reampling_BA}
S = \argmin_{E} \lambda_1 S_I + \lambda_2 S_M,
\end{dmath}
where $\lambda_1$ and $\lambda_2$ are the weighting scalars. While applying resampling to the incremental reconstruction loop can improve the 3D trajectories and the temporal alignment, it requires inverting a large and dense matrix of the DCT coefficients, which is computationally expensive. Thus, we only use this resampling scheme as a post-processing.

\subsection{Evaluation on Motion Capture Data}

\begin{figure}[t]
\includegraphics[width=\linewidth]{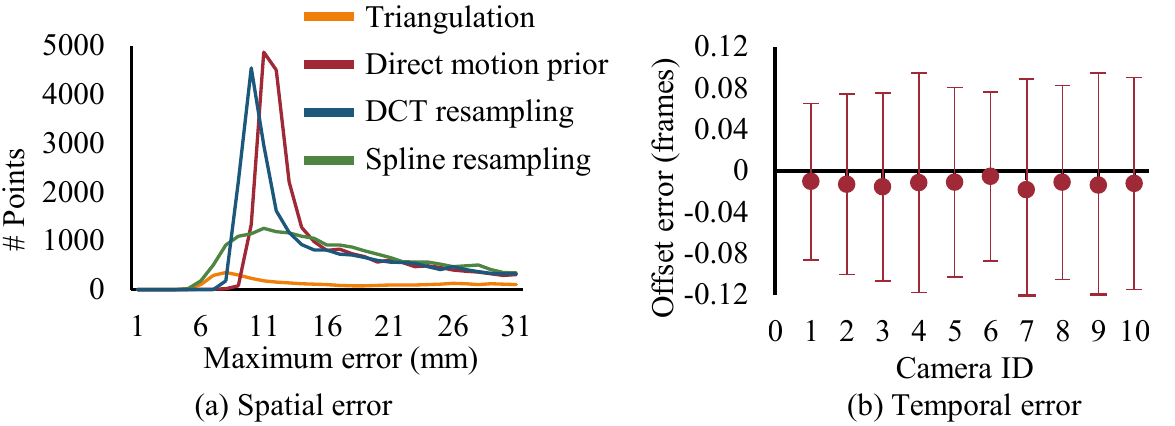}
\caption{Evaluation of the motion priors on the Motion Capture database for simultaneous 3D reconstruction and sub-frame temporal alignment. (a) Spatially, the trajectories estimated using the motion prior achieves higher accuracy than generic B-spline trajectories basis. Frame level alignment geometric triangulation spreads the error to all cameras and estimates less accurate 3D trajectories. (b) Temporally, our motion prior based method estimates the time offset between cameras with sub-frame accuracy. }
\label{fig:Mocap2D_4D}
\end{figure}

We validate the proposed spatiotemporal bundle adjustment on synthetic data generated from the CMU Motion Capture database~\cite{CMU-Mocap}. We sequentially distribute the ground truth trajectory, captured at 120fps, to 10 global shutter perspective cameras with a resolution of 1920x1080 and 12fps. All cameras are uniformly arranged in a circle and capturing the scene from 3 m away. We randomly add 3000 background points arranged in a cylinder of radius 15 m centered at dynamic points. The relative offsets, discretized at 0.1 frames, are randomly varying for every sequence. None of the offsets generates cameras observing the 3D points synchronously. We assume that the initial offsets are within 2-3 frames accurate, which is the case for most geometry-based alignment methods. We also add zero mean Gaussian noise of 2 pixels standard deviation to the observed 2D trajectories. For the divide and conquer approach, we split the camera into three groups of 4 nearby cameras each.

\begin{table}[]
\centering
\label{Tab1}
\begin{tabular}{c|c|c|c|c|c|c|}
\cline{2-7}
& \multicolumn{4}{c|}{Incremental}              & \multicolumn{2}{c|}{Divide and conquer} \\ \cline{2-7} 
& Geometry & Spline & MP & R & MP     & R     \\ \hline
\multicolumn{1}{|c|}{Static}  & 3.45     & 2.93   & \textbf{2.54} & \textbf{2.41}       & 2.56     & \textbf{2.41}          \\ \hline
\multicolumn{1}{|c|}{Dynamic} & 17.8     & 1.68   & \textbf{0.85} & \textbf{0.74}       & 0.89     & 0.75           \\ \hline
\end{tabular}
\small{\caption{The reprojection error for the entire CMU Mocap dataset. Motion prior (MP) refers to the results for all cameras without the DCT resampling step (R).}}
\end{table}

The reconstruction and alignment errors are summarized in Fig.\ref{fig:Mocap2D_4D} and Table 1. Spatially, the point triangulation of the frame-accurate alignment propagates the error to all cameras and gives the worst result. Trajectories reconstructed using generic interpolation functions such 3D cubic B-spline basis give much smaller error than the point triangulation. However, it also arbitrarily smooths out the trajectories and is inferior to our method. While both the direct motion prior and DCT resampling have similar mean error (direct: 6.6 cm, DCT: 6.5 cm), the former has a larger maximum error due to the noise in approximating the velocity. Temporally, our method can estimate ground truth offset at sub-frame accuracy with low uncertainty. The divide and conquer approach is quantitatively as accurate as the incremental reconstruction and alignment method while being approximately 30 times faster. We also observe that this approach produces no temporal inconsistency between camera groups for all trials.   

\section{Dynamic Reconstruction of Human Body}
While the spatiotemporal bundle adjustment can accurately estimate the 3D trajectory of dynamic points, such trajectories are often sparse are insufficient to fully visualize the scene content. In this section, we leverage recent advances in visual semantic understanding to fit a statistical human body model to the observed semantic cues (i.e., 2D body part segmentation and anatomical keypoints), along with the sparse 3D trajectories recovered using sparse spatiotemporal bundle adjustment. \Blue{Our key is to use the spatiotemporal calibration for accurate frame sequencing and the motion prior to link single-frame estimations over the entire multiview sequences for 3D human body fitting.} Due to the imprecision in localizing the semantic cues, we fix camera spatiotemporal parameters and only optimize for the human body. 

\subsection{Statistical Body Model} We use the SMPL model~\cite{loper2015smpl}, a linear blend shape model of body shape that can be deformed via linear blend skinning, to present the human body. This model $V(\Omega, \Phi, \Gamma)$ is a triangle mesh, composed of 6890 vertices, and is parameterized by gender, 10 identify shape coefficients $\Omega$, 24 3-DoF joints $\Phi$, and a root translation $\Gamma$. The 3D location of the body joints $X$ corresponding to a particular pose configuration is given by the joint regressor matrix $R$, a sparse matrix modeling the influence of small number of surface vertices around the joint, $X = R V(\Omega, \Phi, \Gamma)$. In our case, $R$ is slightly different from~\cite{loper2015smpl} as our joints are defined according to the OpenPose keypoints format~\cite{cao2018openpose}.

\subsection{Body Alignment Objective}
We fit the SMPL model to the observed semantic body part, the keypoints, and the sparse 3D trajectory by optimizing the following cost:
\begin{dmath}
\label{eq:BodyAllCost}
S = \argmin_{\Omega, \mathbf{\Phi}(t),\mathbf{\Gamma}(t)}\lambda_{S_J}S_{J} + \lambda_{S_T}S_{T} + \lambda_{S_S}S_{S}+ \lambda_{S_M}S_{M}+ \lambda_{S_B}S_{B}, 
\end{dmath}
where $S_J, S_S$ are the image evidence cost, capturing the body joint and part silhouette, respectively, $S_T$ is the cost induced by the sparse 3D trajectory on the body, $S_M$ is the least kinetic motion prior loss imposed on the 3D body joints, and $S_B$ is the body pose and shape prior cost. We normalize each of these costs by the number of their contributing residuals before applying the weights $\lambda_{S_K}, \lambda_{S_T}, \lambda_{S_S}, \lambda_{S_M}, \lambda_{S_B}$ to each cost. \Blue{Several of these costs are common ($S_J,S_S,S_M$~\cite{bogo2016keep,lassner2017unite,huang2017towards}, $S_B$~\cite{joo2018total}) and their effectiveness have been shown in prior work under various forms. We list the definition of the more standard losses in Tab.~\ref{tab:3D_Body_standard} and describe our customized losses in details below.}\\

\begin{table*}
\centering	
\begin{tabular}{cc}
\begin{minipage}{0.5\textwidth}
\begin{flushleft}
\begin{tabular}{|l|l|}
\hline
$S_J$        & $\sum_{c=1}^{C} \sum_{f=1}^{F_c} \sum_{j=1}^{J}I_c^j(f)\sigma^j_c(f)\bigg(\frac{||\pi_c(f, X^j) -x_c^j(f)||}{\sigma_J} \bigg)^2$  \\ \hline

$S_M$ & $\sum_{t \in T} \sum_{X_j} \Big\Vert\frac{X^j(t^(i+1)) - X^j(t^i)}{t^{i+1}-t^i+\epsilon}\Big\Vert^2 \frac{(t^{i+1}-t^i)}{\sigma_{M_j}}$ \\ \hline

$S_B$              & $\lambda_{\Omega}\mathcal{N}(\Omega) + \lambda_{\Phi} \sum_{c=1}^C \sum_{f=1}^{F_c} \mathcal{N}(\Phi(f))$ \\ \hline
\end{tabular}\\	
\end{flushleft}
\end{minipage}&

\begin{minipage}{0.5\textwidth}{
$C$: number of cameras\\
$F$: number of frames\\ 
$J$: number of 2D keypoints detected by OpenPose~\cite{cao2018openpose}\\
$T$: number of observed time samples for the entire event\\
$X_j$: 3D body joint $j$\\
$I_c^{j}(f)$: visibility indicator\\ 
$\mathcal{N}$: zero mean standard normal distribution\\ 
$\sigma_J$: variation in 2D detection\\
$\sigma_{M_j}$: variation in bone length\\
}
\end{minipage}
\end{tabular}
\caption{Common 3D body fitting losses used in Eq.~\ref{eq:BodyAllCost}. Please refer to the text for other losses.}
\label{tab:3D_Body_standard}
\end{table*}

\boldstart{Sparse 3D trajectory constraints}: This cost function penalizes variance of the distances $L(.,.)$ between the point $X^n$ along the 3D trajectories to the two nearest joints within the same body part, $X^j \in 2NN(X^n)$. Our loss is expressed as
\begin{align}
\label{eq:3DTrajectory}
S_{T} = \sum_{c=1}^{C} \sum_{f=1}^{F_c} \sum_{n=1}^{N} \sum_{X^j}  I_c^n(f) \bigg(\frac{||L(X^j,X^n) - \overline{L}(X^j,X^n)||}{\sigma_T}\bigg)^2, 
\end{align}
where $\sigma_T$ is a scalar capturing the uncertainty of the dynamic 3D point, $I_c^n(f)$ is a binary showing the availability of $X^n(t)$ on to camera $c$ at frame $f$, and $l$ is the Euclidean distance between two points. We add the extra variable $\overline{L(.,.)}$ as the average distance between points over the entire course of motion, to the optimization. Despite the non-rigid body and cloth deformation, we expect the deviation of the instantaneous point distance to be close to its average over time. Empirically, we observe that this loss function improves tracking robustness especially in rare cases of simultaneous multiview semantic detector failure when body parts corresponding to different people are grouped together.  

\boldstart{Body-part alignment cost}: \Blue{This cost function improves the alignment between the projected body vertices with the detected body part segmentation and discourages any projected body vertices not contained inside the segmentation. Our loss is formulated as
\begin{align}
\label{eq:Vis}
S_{S} = \sum_{c=1}^{C} \sum_{f=1}^{F_c}\sum_{p=1}^P\sum_{v\in V_p(\Omega,\Phi(t),\Gamma(f))}  I_c^p\rho\bigg(\frac{DT_c^p(f)(\pi_c(f,v))}{\sigma_P}\bigg),
\end{align}
where $DT_c^p(f)$ is the distance transform of the $p$ body part segmentation, estimated using DensePose~\cite{alp2018densepose}, in camera $c$ at frame $f$, $\rho$ is the German-McClure robust loss to account for possible segmentation inaccuracies, $\sigma_P$ is a scalar approximating the uncertainty in estimating the part segmentation, and $I_c^p$ is the binary visibility of the part segmentation. $P$ includes both the body part for fine-grain fitting and their union (silhouette) for free-space reasoning. While the silhouette is particularly useful for occluded body parts in self-occlusion scenario (e.g. the arm behind the body case where the segmentation of the visible parts are mostly clean), for inter-person occlusion, it adversely encourages the missing parts to go inside the truncated silhouette of the occluded person. We handle this scenario by detecting the occluded person by considering the depth ordering of the initial mesh and disable the term.}

\subsection{Optimization Strategy}
Due to the complexity of the human body pose, a direct optimization of Eq.~\ref{eq:BodyAllCost} converges slowly and often fails to produce accurate body fitting. We solve the problem in three stages. (1) full sequence spatiotemporally coherent 3D human skeleton estimation(2), per-time-instance human model fitting to the skeleton, and (3) window-based accurate and temporally coherent body pose and shape fitting. \Blue{Empirically, since the potential missing skeleton joints in stage 1 is spatiotemporally resolved, the per-frame body fitting in stage 2 is naturally more accurate, which further speeds up the convergence in stage 3.} These stages are described below.

\boldstart{Stage 1: Coherent 3D human skeleton estimation} For each person in the scene, we wish to estimate a temporally and physically consistent human skeleton model for the entire sequence. This is done by minimizing an energy function consisting of $S_J$, $S_M$, and the prior on human skeleton which is formulated as
\begin{align}
\label{eq:SkeletonPrior}
S_b = \sum_{t\in T} \sum_{q\in Q} \bigg(\frac{||L(q, t)-\overline{L}(q)||}{\sigma_L}\bigg)^2, \\
S_{lr} = \sum_{t\in T} \sum_{(l,r)\in S} \bigg(\frac{||L(l,t) - L(r,t)||}{\sigma_S}\bigg)^2, 
\end{align}
where $Q$ is the set of keypoint connectivity within all rigid body parts, $S$ denotes the set of joints of the corresponding left and right limb, $\{\sigma_L, \sigma_S\}$ captures the precision of the symmetry and left-right consistency constraints. These priors enforce the left-right symmetry of the body bone length and penalize large changes between the bone length estimated each time instance and average bone length $\overline{L}$ over the entire sequence. As in Sec. 3, the initial 3D skeleton is obtained by geometric triangulation. We weigh the costs equally and optimize them for the entire sequence at once. Lastly, we resample the 3D joint along the temporal axis using DCT to fill in the missing skeleton due to occlusion.

\boldstart{Stage 2: Per-instance human model fitting} Given the coherent 3D skeleton at all time instances, we fit the SMPL model to the skeleton independently at each time instance in order to gain resilience to fitting failure. This is done by optimizing a cost function consisting of the 3D-3D SMPL joint to our skeleton cost, and body shape and pose prior $S_B$. The 3D-3D cost is written as
\begin{align}
\label{eq:3D-3Djoints}
S_{3D-3D} = \sum_{t\in T} \sum_{j=1}^{J}\bigg(\frac{||X^j - \tilde{X}j||}{\sigma_{3D}}\bigg)^2, 
\end{align}
where $\tilde{X}$ is the estimated skeleton at stage 1 and $\sigma_{3D}$ is the expected noise in 3D estimation. Empirically, we found this approach is fast and gives good approximation for the last stage.

\boldstart{Stage 3: Window-based human model fitting} We optimize Eq.~\ref{eq:BodyAllCost} for the SMPL body shape and pose in overlapping windows. For the overlapping region, we fix the optimized parameters to those of the previous windows to ensure consistent body shape estimation.

\subsection{Evaluation on Motion Capture Data}
\begin{figure*}[t]
\includegraphics[width=\linewidth]{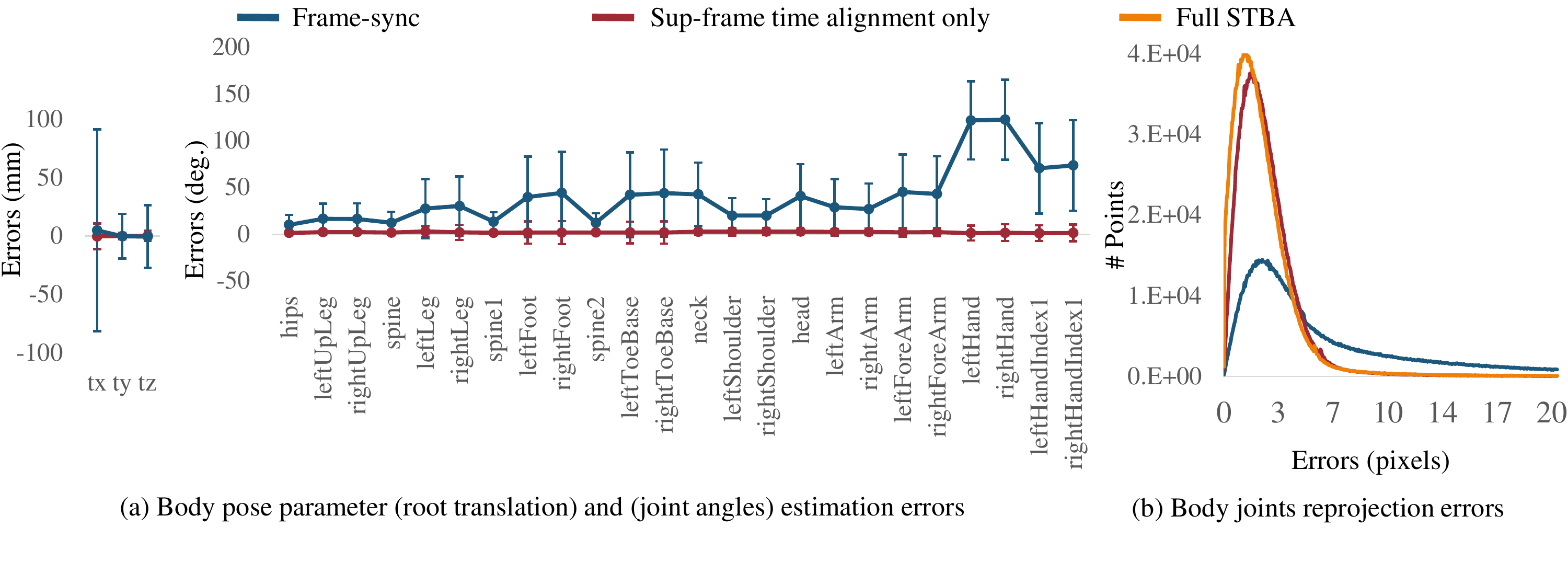}
\caption{\Blue{The effect of the spatiotemporal bundle adjustment on human body pose estimation on the CMU portion of AMASS dataset~\cite{AMASS} using (a) the body pose parameters and (b) the reprojection error of the body joints as metrics. The spatiotemporal calibration parameters leads to significantly lower body pose estimation errors than the synchronized captured assumption. The difference between using only the sub-frame temporal offsets and full spatiotemporal calibration parameters are less noticeable especially for the body pose estimation.}}
\label{fig:Mosh_STBA}
\end{figure*}

\Blue{This section analyzes the effect of the spatiotemporal bundle adjustment on the SMPL pose parameters estimation (i.e., root translation $\Gamma$ and joint rotation angles $\Phi$) on the data generated from the CMU portion of the AMASS dataset~\cite{AMASS}. Our synthetic experiment setup is the same as in Sec. 3.4 except that the inputs are the noisy 2D projections of the 3D SMPL body keypoints to the cameras. The body keypoints include the original SMPL joints~\cite{loper2015smpl}, the tip of the thumb and the middle finger, and the OpenPose foot keypoints~\cite{cao2018openpose}. The 2D noise is Gaussian noise with zero mean and 2 pixels standard deviation. The relative offsets and the camera poses are estimated using the spatiotemporal bundle adjustment described in Sec. 3. We use the SMPL pose parameter errors and the reprojection errors of estimated body joints as metrics and compare the fitting results on three cases: synchronized capture with triangulation constraints~\cite{joo2018total,huang2017towards}, unsynchronized capture using the estimated time offsets only, and unsynchronized capture using the time offsets and camera poses from spatiotemporal bundle adjustment (full STBA). Note that we only use the keypoints and assume known body shape $\Omega$ in this experiment for simplicity. We believe that this is sufficient to analyze the effect of the spatiotemporal bundle adjustment}

\Blue{We summarize the results in Fig.~\ref{fig:Mosh_STBA}. Evidently, the spatiotemporal calibration parameters lead to significantly better body fitting than the synchronized frame capture assumption. As shown in Fig.~\ref{fig:Mosh_STBA}b, the biggest influencing factor is to properly model the time offsets between cameras. Using the full spatiotemporal bundle adjustment results leads to marginal fitting improvement. We also observe that the mean angle error for all joints in our approach are within 4 degrees, which complies with the state of art marker-based motion capture~\cite{kessler2019direct}.}

\section{Analysis on Real Handheld Camera Data}

\begin{figure}[t]
\includegraphics[width=\linewidth]{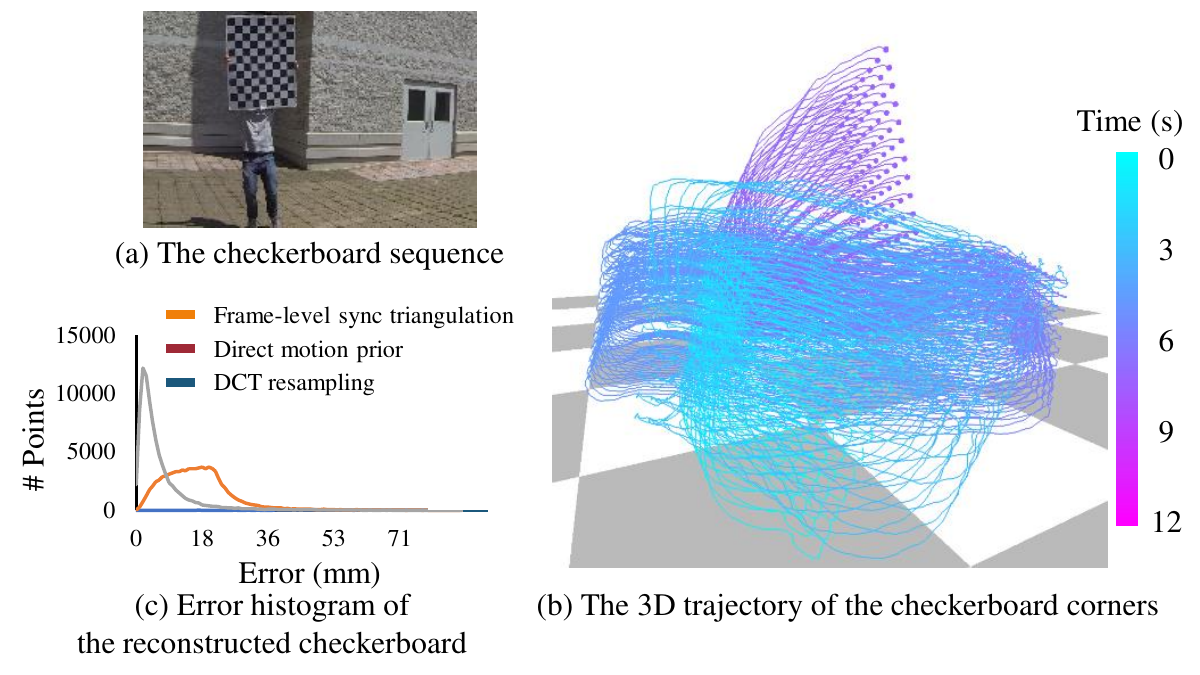}
\caption{Accuracy evaluation of the checkerboard corner 3D trajectories. While the reconstruction is conducted independently at every corner, the estimated 3D trajectories assemble themselves in the grid-like configuration. Our methods produce trajectories with significantly smaller error than naive geometric triangulation.}
\label{fig:Checker_acc}
\vspace{-.5cm}
\end{figure}

\begin{figure*}[t]
\includegraphics[width=\linewidth]{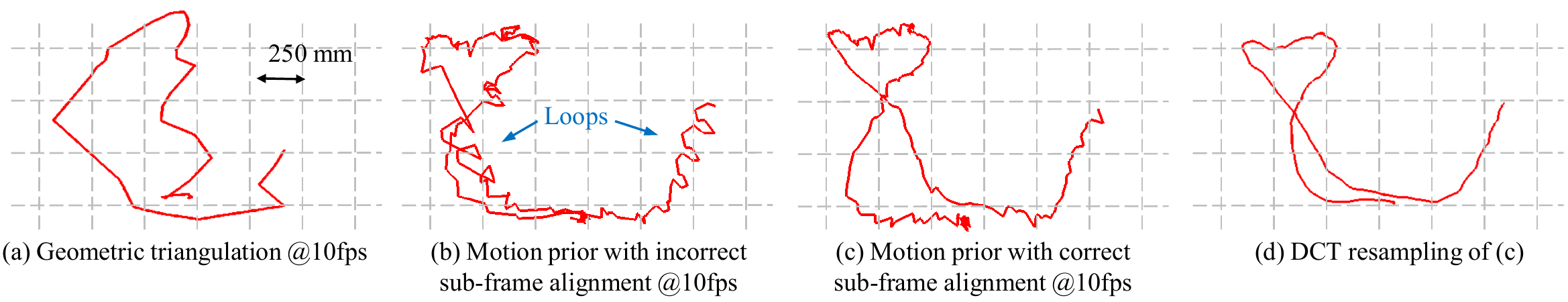}
\centering
\caption{Effect of accurate subframe alignment for the 3D trajectory estimation. (a) Point triangulation of frame accurate alignment gives large reconstruction error and creates different 3D shapes with respect to other methods. (b) Incorrect sub-frame alignment generates 3D trajectory with many loops. (c) Trajectory estimated from correct sub-frame alignment is free from the loops. (d) Using DCT resampling for (c) gives smooth and shape preserving 3D trajectory.}
\label{fig:Checker_Subframe}
\vspace{-0.1in}
\end{figure*}

\subsection{Data Preprocessing}
We create a pipeline that takes video streams for multiple temporally unaligned and spatially uncalibrated cameras and produces the spatiotemporal calibration parameters as well as the 3D static points and dynamic trajectories. We show the results for 3 scenes: checkerboard, jump, and dance, captured by either smartphone or GoPro Hero 3 cameras, all of which are rolling shutter cameras. We quantify the error in 3D trajectory estimation and effect of sub-frame alignment using the Checkerboard sequence, captured by 7 Gopro at 1920$\times$1080 and 60fps for 20s. The Jump sequence, captured by 8 Gopro at 1280$\times$720 at 120fps for 4s, serves to demonstrate our ability to handle fast motion using low frame rate cameras. The Dance scene, captured by 6 iPhone 6 and 6 Samsung Galaxy 6 at 1920$\times$1080 and 60fps for 20s, showcases the situation where the static background and dynamic foreground are separated by a large distance. \Blue{For sequences captured by smartphone cameras, we solve for per-frame camera intrinsics because of their auto-focus function.} For all scenes, we purposely down-sample the frame rate to simulate faster motion, which invalidates the geometry constraints for unsynchronized cameras and stresses the essence of the motion prior: 10fps, 30fps, and 15fps for Checkerboard, Jump, and Dance, respectively. \Blue{In practice, we expect reconstructing such fast motion is more challenging due to motion blur which significantly degrades the 2D correspondences.}

\boldstart{3D corpus and initial camera pose estimation:} We track SIFT features using affine optical flow~\cite{baker2004lucas} and sample keyframes, defined as frames where the number of tracked features drop $40\%$ from the last keyframe, from all videos. These keyframes are passed to a SfM  pipeline~\cite{VisualSfm,schonberger2016structure} to build the 3D corpus of the scene. We register other frames to this corpus using the r6P algorithm and refine their parameters using the Cayley transform model~\cite{alblr6p}. No temporal regularization is performed during the registration to preserve the abrupt a camera motion frequently observed due to the camera holder's footstep. 

\boldstart{Rolling shutter scanning speed estimation:} Consider a moving camera observing static features. Geometrically, this camera can also be viewed as being static and observing moving features. We estimate the camera rolling shutter readout speed by assuming the 3D location of these moving features also obeys the least kinetic motion prior for the duration of 1 frame. Denote ${X_v(f),X_v(f+1)}$ as the virtual location of the static feature captured exactly at the first row of of frame ${f, f+1}$, respectively, and $X_(t)$ is the true location of the same feature observed in the image, $t\in [t(f), t(f+1)]$. Under the least kinetic assumption (e.g., constant velocity) and vertical rolling shutter readout, we can present the 3D location of the observed feature as 
\begin{align}
\label{eq:readout}
X(t) = X_v(f) + \gamma \frac{r}{h} ( X_v(f+1)- X_v(f)), 
\end{align}
where $r$ is the image row of the observed feature and $h$ is the image height. Using this representation of $X$ to optimize Eq.~\ref{eq:ImgCost}, we can estimate the rolling shutter readout speed $\gamma$ for each camera. \Blue{Our approach estimates consistent $\gamma$ for cameras of the same model in our three testing sequences.}

\boldstart{Corresponding 2D trajectory generation:} We detect and match SIFT features across cameras at evenly distributed time instances. We discard matches with low gradient score and track the remaining points both forward and backward in time using affine template matching. The backward-forward consistency check is used to discard erroneous optical flow during tracking. Finally, we check for the appearance consistency between patches of the first and the last frame using Normalized Cross Correlation and remove the entire trajectory if the score is below 0.8.

\noindent\textbf{Trajectory classification:} We exploit the fact that triangulation based methods work for static points but produce large errors for dynamic points in order to identify 2D trajectories of dynamic points. This is done using these two heuristics: (1) the reprojection error of a static point should be small regardless of which camera frame it is triangulated from. We randomly sample frames along the 2D trajectory to triangulate and consider the 2D trajectory as belonging to a static point if the reprojection threshold is smaller than 3 pixels for more than 80\% of the time. (2) the reprojection error of a dynamic point forms a steep valley as the time offset passes by its true value. We reject any set of trajectories as belonging to a dynamic point if the minimum of the cost valley is not smaller than 80\% of the average cost. 

\subsection{Sparse spatiotemporal bundle adjustment}

\begin{figure}[t]
\includegraphics[width=\linewidth]{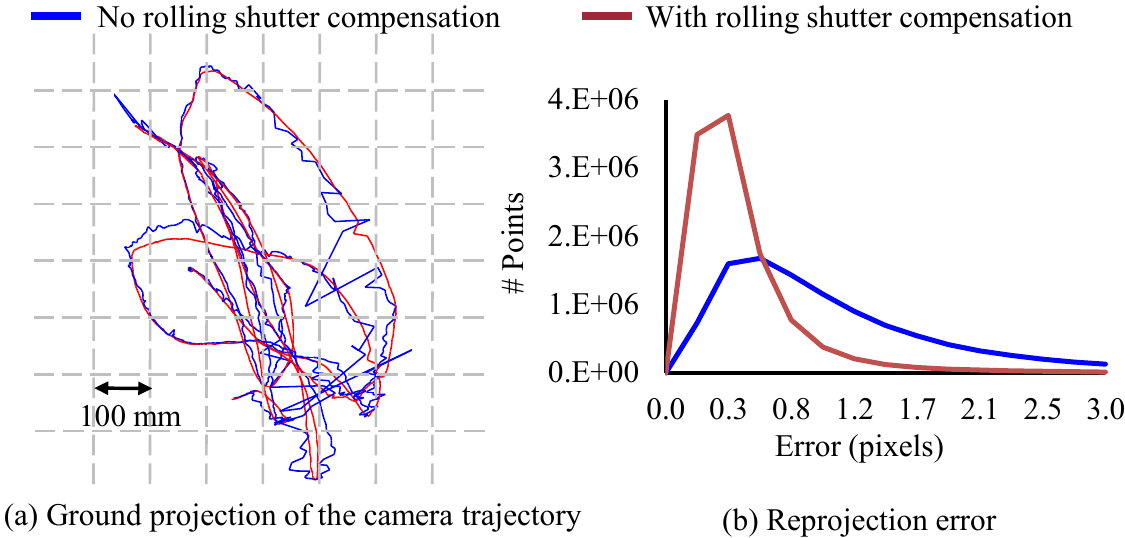}
\centering
\caption{Analysis of the spatial camera calibration for the Checkerboard sequence with different camera models.}
\label{fig:RollingShutter_CheckerS}
\end{figure}

\begin{table}[t]
\centering
\label{tab:gamma}
\begin{tabular}{c|c|c|c|c|}
\cline{2-5}
& \multicolumn{2}{c|}{\begin{tabular}[c]{@{}c@{}}No $\gamma$ \\ Static-Dynamic\end{tabular}} & \multicolumn{2}{c|}{\begin{tabular}[c]{@{}c@{}}With $\gamma$\\ Static-Dynamic\end{tabular}} \\ \hline
\multicolumn{1}{|c|}{Checkerboard} & 0.70                                    & 1.52                                    & 0.67                                     & 1.21                                    \\ \hline
\multicolumn{1}{|c|}{Jump}         & 0.61                                    & 1.55                                    & 0.59                                     & 1.34                                    \\ \hline
\multicolumn{1}{|c|}{Dance}        & 0.85                                    & 2.38                                    & 0.82                                     & 2.13                                    \\ \hline
\end{tabular}
\caption{The effect of modeling the rolling shutter readout on the reconstruction accuracy. While the temporal sequencing between cameras is still correct (due to the artificial down-sampling of the frame rate), modeling $\gamma$ results in smaller the reprojection error. Inaccurately reconstructed dynamic points also (slightly) negatively affect the reconstruction of static points.}
\vspace{-.5cm}
\end{table}

\begin{table*}[t]
\centering
\label{Tab2}
\begin{tabular}{c|c|c|c|c|c|c|c|c|c|c|c|c|c|c|}
\cline{2-15}
& \multicolumn{4}{c|}{\scriptsize Geometry} & \multicolumn{10}{c|}{\scriptsize Motion prior}  \\ \cline{2-15} 
& \scriptsize \#Trajectory & \begin{tabular}[c]{@{}c@{}}\scriptsize Avg samples\\ \scriptsize per trajectory\end{tabular} & \multicolumn{2}{c|}{\begin{tabular}[c]{@{}c@{}}\scriptsize RMSE (pixels)\\ \scriptsize Static --- Dynamic\end{tabular}} & \scriptsize \#Trajectory & \begin{tabular}[c]{@{}c@{}}\scriptsize Avg samples \\ \scriptsize per trajectory\end{tabular} & \multicolumn{2}{c|}{\begin{tabular}[c]{@{}c@{}}\scriptsize RMSE$^1$ (pixels)\\ \scriptsize Static --- Dynamic\end{tabular}} & \multicolumn{2}{c|}{\begin{tabular}[c]{@{}c@{}}\scriptsize RMSE$^{1*}$ (pixels)\\ \scriptsize Static --- Dynamic\end{tabular}} & \multicolumn{2}{c|}{\begin{tabular}[c]{@{}c@{}}\scriptsize RMSE$^2$ (pixels)\\ \scriptsize Static --- Dyamic\end{tabular}} & \multicolumn{2}{c|}{\begin{tabular}[c]{@{}c@{}}\scriptsize RMSE$^{2*}$ (pixels)\\ \scriptsize Static ---Dynamic\end{tabular}} \\ \hline
\multicolumn{1}{|c|}{\scriptsize Checkerboard} & \scriptsize 88           & \scriptsize 179.8                                                                & \scriptsize 0.67                                           & \scriptsize 6.59                                           & \scriptsize 88           & \scriptsize 1023.0                                                                &\scriptsize  0.67                                           &\scriptsize  1.21                                           & \scriptsize 0.65                                           & \scriptsize 1.15                                           & \scriptsize0.68                                           & \scriptsize 1.2                                           & \scriptsize0.67                                           & \scriptsize 1.16                                          \\ \hline
\multicolumn{1}{|c|}{\scriptsize Jump}         & \scriptsize 717          & \scriptsize 36.4                                                                 & \scriptsize 0.59                                            & \scriptsize1.91                                           & \scriptsize 3231         & \scriptsize 127.8                                                                 & \scriptsize 0.59                                           & \scriptsize 1.34                                           & \scriptsize 0.5                                            & \scriptsize 1.26                                           & \scriptsize 0.59                                           & \scriptsize 1.36                                          & \scriptsize 0.51                                           & \scriptsize 1.26                                          \\ \hline
\multicolumn{1}{|c|}{\scriptsize Dance}        & \scriptsize \scriptsize 577          & \scriptsize22.3                                                                 & \scriptsize 0.82                                           &\scriptsize 5.23                                           & \scriptsize 4105         & \scriptsize 216.4                                                                 & \scriptsize 0.82                                           & \scriptsize \scriptsize 2.12                                           & \scriptsize 0.85                                           & \scriptsize \scriptsize 1.71                                           & \scriptsize 0.83                                           & \scriptsize 2.14                                          & \scriptsize 0.87                                           &\scriptsize  1.72                                          \\ \hline
\end{tabular}
\caption{Reconstruction accuracy comparison between geometric triangulation and our proposed method. RMSE$^1$ and RMSE$2$ are the results obtained by the incremental reconstruction and alignment and divide and conquer approaches, repetitively. The $*$ denotes the results after resampling. Both approaches are equally accurate and the divide and conquer scheme is at least 10 times faster. Please see the text for more details.}
\vspace{-.5cm}
\end{table*}

\begin{figure*}[t]
\includegraphics[width=\linewidth]{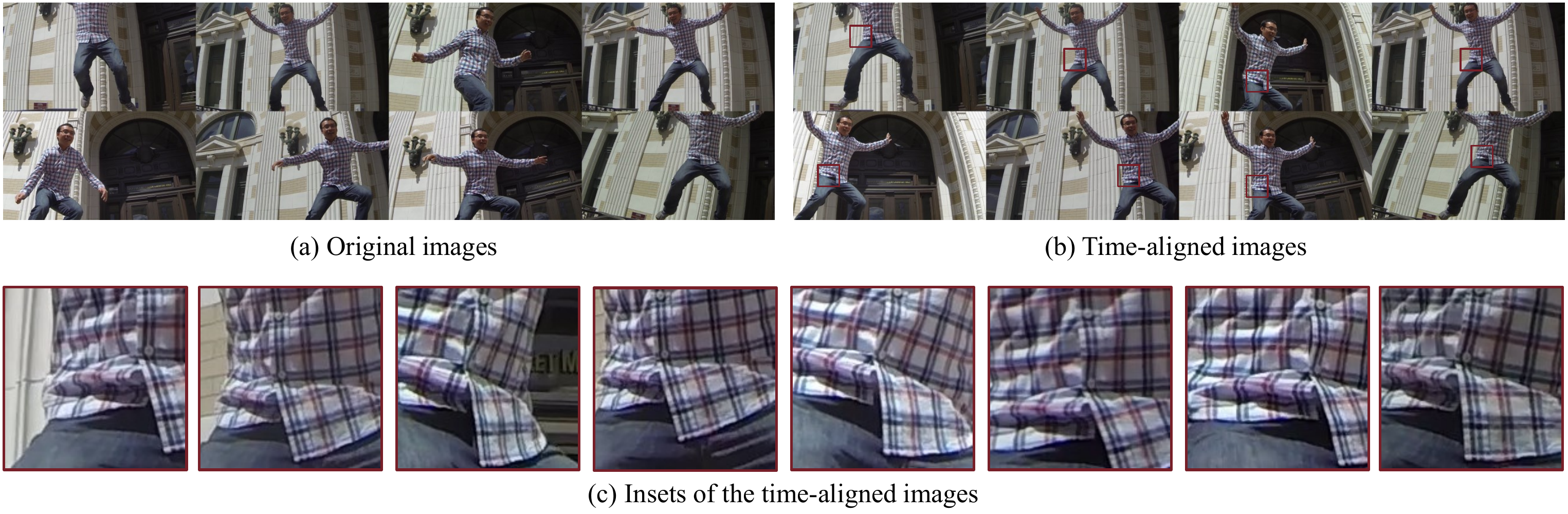}
\centering
\caption{Temporal alignment. (a) Original unaligned images. (b) Our aligned images, estimated from temporally down-sampled video at 30fps, are shown for the original video captured at 120fps. (c) Inset of aligned images. The shadows casted by the folding cloth are well temporally aligned across images.}
\label{fig:Jump_aligned}
\end{figure*}

\begin{figure*}
\includegraphics[width=\linewidth]{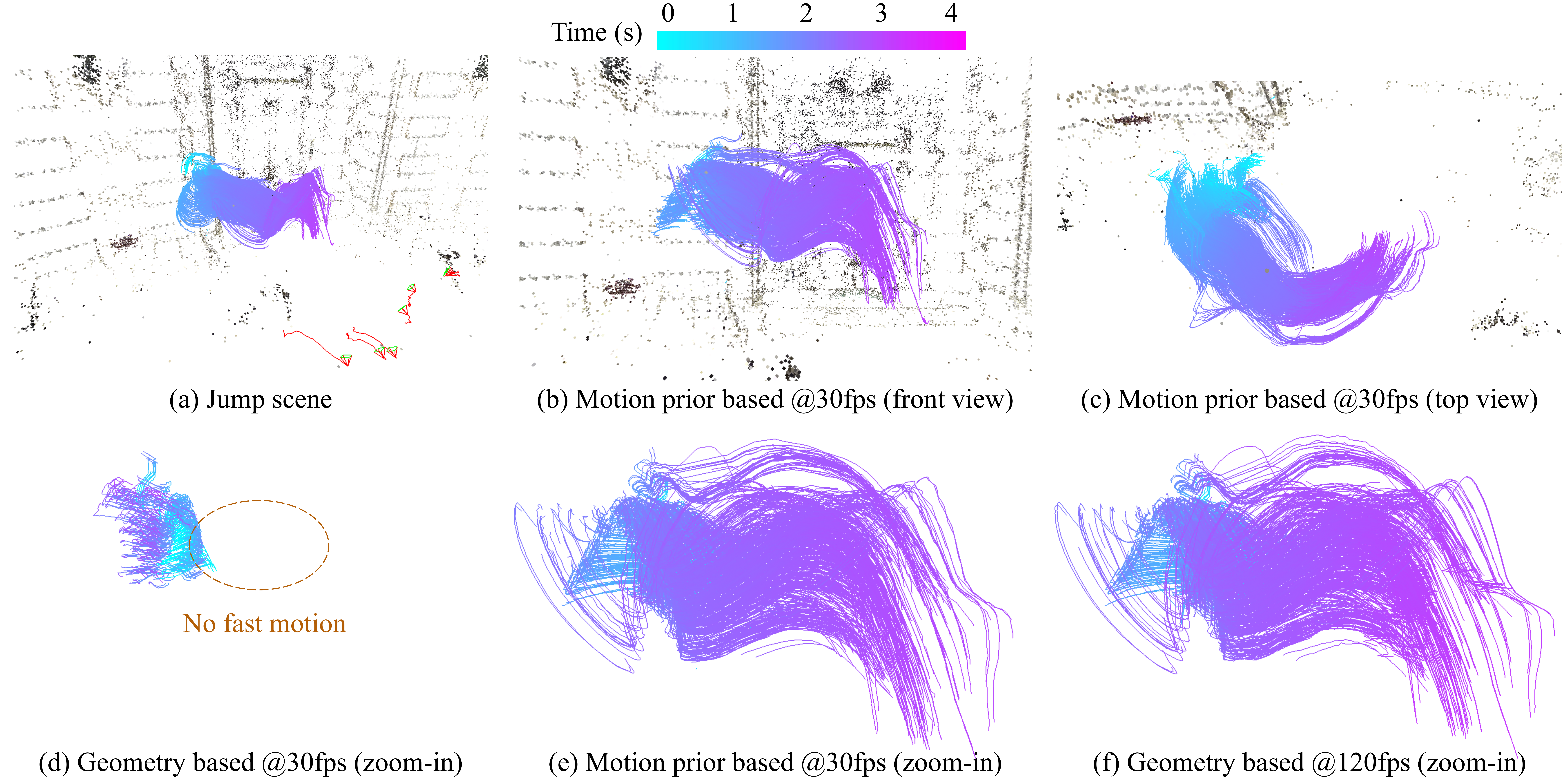}
\centering
\caption{Jump scene. Point triangulation of frame-accurate alignment fails to reconstruct the fast action happening at the end of the sequence. Conversely, our motion prior based approach produces plausible reconstruction for the entire course of the action even with relatively low frame-rate cameras. Trajectories estimated from our approach highly resemble those generated by the frame-accurate alignment and triangulation at 120fps.}
\vspace{-.5cm}
\label{fig:Jump_Traj}
\end{figure*}

\begin{figure*}
\includegraphics[width=0.975\linewidth]{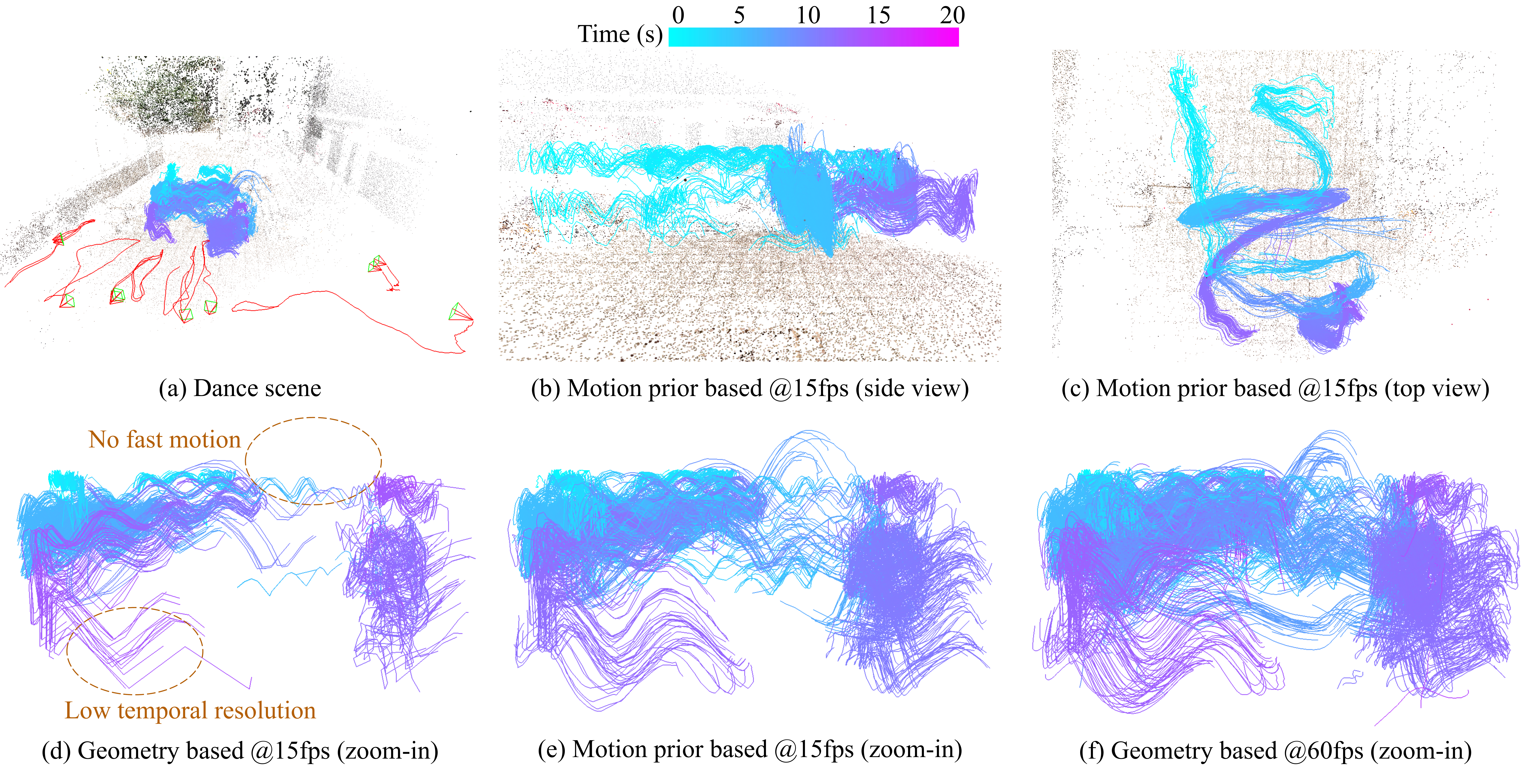}
\centering
\caption{Dance scene. The 3D trajectories are estimated using ten 15fps cameras. Noticeably, the trajectories generated from frame accurate alignment and triangulation are fewer, shorter, and have lower temporal resolution than those reconstructed from motion prior based approaches.}
\label{fig:Dance_Traj}
\end{figure*}

\begin{figure*}
\includegraphics[width=0.975\linewidth]{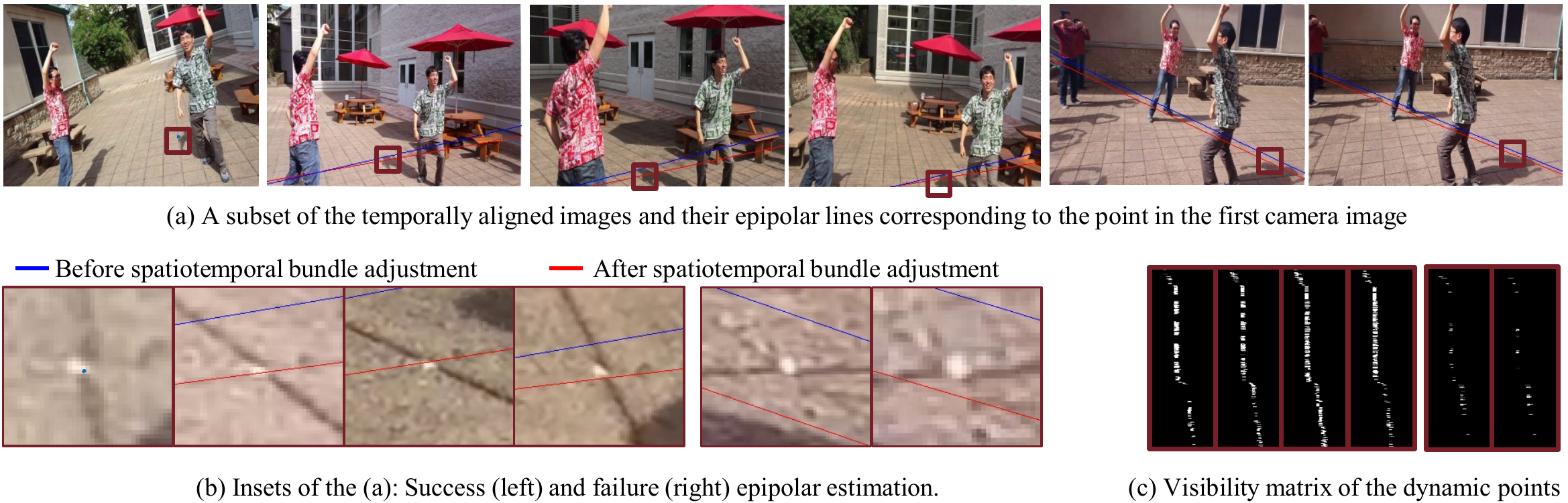}
\centering
\caption{Evaluation of the spatiotemporal calibration. The blue and red lines are the estimated epipolar lines before and after spatiotemporal bundle adjustment, respectively. The epipolar lines estimated after spatiotemporal bundle adjustment have noticeable improvement at the foreground for cameras with a large number of visible dynamic points.}
\label{fig:Dance_STBA}
\end{figure*}

\begin{figure*}
\includegraphics[width=\linewidth]{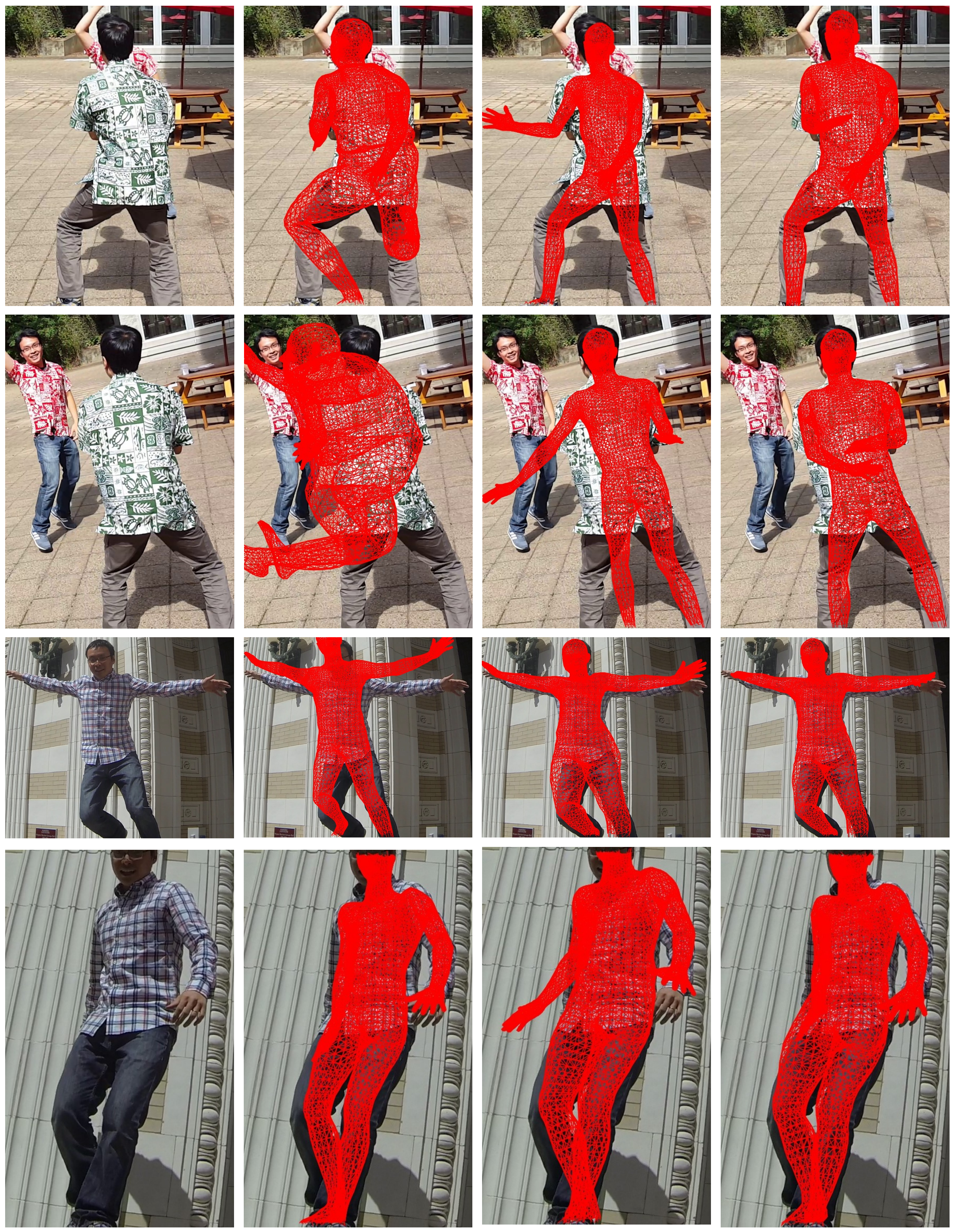}
\centering
\caption{The effect of subframe alignment for human shape fitting for the Dance and Jump scenes. Given the input frames (first column), due to fast motion, shape fitting assuming frame-level alignment and geometric constraint (second column) fails to match the silhouette of the person. Our motion prior based shape fitting to the 2D body joints (third column) better aligns the observed silhouette but fails when the joints are not visible. Adding the silhouette constraint to the motion prior based fitting produces the best results (last column).}
\label{fig:Fitting_ablative}
\vspace{-0.1in}
\end{figure*}

\begin{figure*}
\includegraphics[width=\linewidth]{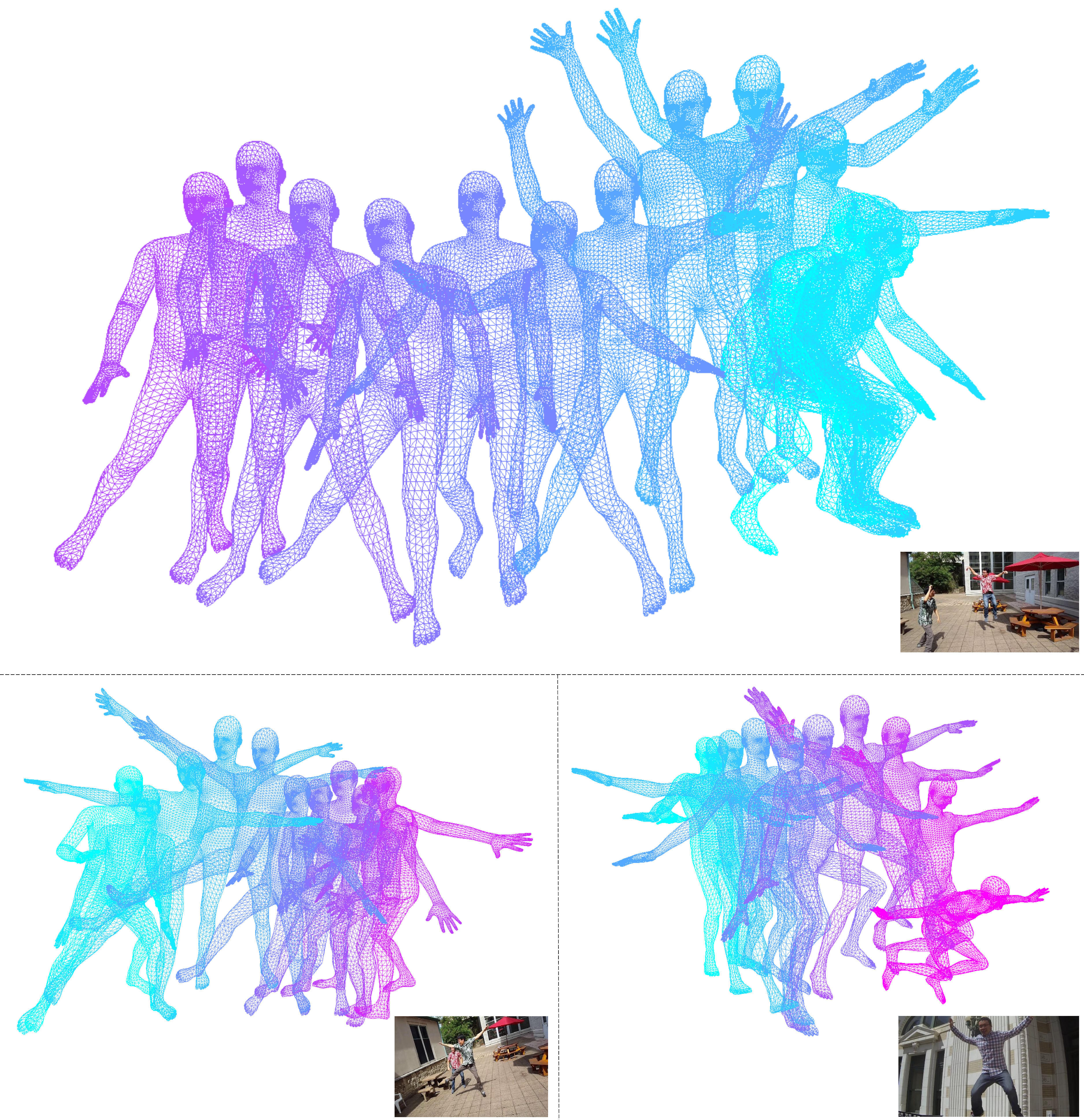}
\centering
\caption{Human mesh fitting for different people in the Dance sequence (top and bottom left) and Jump sequence (bottom right). The color encodes the relative action time. \Blue{Please refer to the supplementary video for the reconstructed body motion.}}
\label{fig:humanMesh}
\vspace{-0.1in}
\end{figure*}

We first evaluate the effect of properly modeling the rolling shutter pose and its readout speed to the reconstruction. As shown in Fig.~\ref{fig:RollingShutter_CheckerS}, spatial modeling the rolling shutter produces significantly more stable camera trajectory and lower reprojection error. \Blue{Interestingly, for artificially down-sampled frame rate videos, since the effective frame duration is increased, the system is less sensitive to possible temporal sequencing error and thus, make accurate estimation $\gamma$ less significant. Yet, as shown in Tab. 2, the reconstructions with $\gamma$ modeled are consistently more accurate with proper $\gamma$ estimation.} Theoretically, while the reconstruction of static points is not affected by $\gamma$, inaccurate reconstruction of dynamic points negatively affects the camera calibration parameters, which in turn affects the static points. For all results represented below, both the rolling shutter camera pose and readout speed are employed.

Tab.3 shows our quantitative evaluation on three video sequences using (a) the re-projection error in pixels for both stationary and dynamic points, (b) the number and average length (time) of the 3D trajectories created using points from multiple views as metrics. Points with reprojection error exceeding a threshold are discarded. Noticeably, our proposed method produces several fold more trajectories, longer average trajectory length, and less reprojection error than geometry approach. For the checkerboard sequence, since the correspondences are known, its 3D points are intentionally not discarded. The resampling scheme consistently further reduces re-projection error for all scenes. \Blue{Similar to the analysis on synthetic data, the divide and conquer scheme is as accurate as the incremental reconstruction and alignment approach but is at least 10 times faster (Checkerboard 10x, Jump 17x, and Dance 40x). The Dance sequence has the largest reduction in processing time because it has the largest number of cameras.}

\boldstart{Checkerboard scene:} Since the ground truth location of the board is unknown, we quantify the reconstruction accuracy by measuring the deviation from the planar configuration for all the checkerboard corners. We reconstruct each corner independently. As depicted in Fig.~\ref{fig:Checker_acc}, the reconstruction using geometric triangulation is at least 80 mm inaccurate. Conversely, most 3D corners estimated from our method have much smaller error (direct motion prior: 33 mm, DCT: 15 mm). Visually, the estimated trajectories using the method assemble themselves in the grid-like configuration of the physical board (see the supplementary video).  

Fig.~\ref{fig:Checker_Subframe} shows the effect of accurate subframe alignment on the trajectory reconstruction. Due to the fast motion, geometry-based method produces trajectories with much different shape than the motion prior based method. We artificially alter the sub-frame of the offsets to create wrong frame sequencing between different cameras and optimize Eq.~\ref{eq:STCost} for the trajectory. This results in trajectories with many small loops, a strong cue of incorrect alignment. Conversely, our reconstruction with correct time alignment is free from the loops. Our final result, obtained by DCT resampling, gives smooth and shape-preserving trajectories. 

\noindent\textbf{Jump scene:} To visually evaluate the alignment, we scale the offsets estimated from the down-sampled videos (30fps) to show the alignment on the original footage at 120fps (see Fig.~\ref{fig:Jump_aligned}). Notice that the shadows cast by the folding cloth are well aligned across images. 
Fig.~\ref{fig:Jump_Traj} shows our estimated trajectories for all methods. The point triangulation of frame-accurate alignment fails to reconstruct the fast action happening at the end of the action. Conversely, our method produces plausible metric reconstruction for the entire action even with relatively low frame-rate cameras. Due to the lack of ground truth data, we compare our reconstruction with the point triangulation using 120fps videos, where few differences between the two reconstructions are observed.

\noindent\textbf{Dance scene:} We estimate per-frame camera intrinsic to account for the auto-focus function of smartphone cameras. Fig.~\ref{fig:Dance_Traj} shows our trajectory reconstruction results. Our method reconstructs fast motion trajectories (jumping), longer and higher temporal resolution trajectories than point triangulation results at 15fps. Since we discard many short 2D trajectories (thresholded at 10 samples), we reconstruct fewer 3D trajectories than geometric triangulation at 60fps. However, the overall shape of the trajectories is similar. 

Interestingly, this scene has a large number of static background points. This adversely reduces the spatial calibration accuracy for the foreground points (see Fig.~\ref{fig:Dance_STBA}). Using our algorithm clearly improves the spatial calibration for cameras with enough number of visible dynamic points. 

\subsection{Human body fitting}

Due to the lack of ground truth data, we qualitatively compare the different shape fitting algorithms by its alignment to the observed person silhouette: (1) triangulation-based method and motion-prior based method using the camera spatiotemporal calibration parameters (2) without and (3) with silhouette constraints. \Blue{Our optimization window spans 150 time instances with 30 instances overlapping in between}. As shown in Fig.~\ref{fig:Fitting_ablative}, due to fast human activity and abrupt camera motion, shape fitting using geometric constraint and frame-level alignment either fails to estimate the body shape (first and second rows in the second column) or does not align well with the observed silhouette (third row, second column). \Blue{In this case, the window-based fitting has to simultaneously enforce motion coherency and potentially conflicting 2D cues depending on the human and camera motion, which deteriorates the fitting parameters.} Our motion prior based shape fitting to the detected body keypoints better aligns the mesh to the images but fails when the detected keypoints are occluded (first, second, and fourth rows in the third column). Additionally due to the small misalignment when the SMPL model internal joint location and the detected joints, especially at the hip joints, using the only detected joints produces sub-optimal body shape (third row, third column). Our full cost function considering the motion prior, the body joints, and detected silhouette produces the best fitting results. We notice that contribution of the sparse dynamic 3D points is negligible for both scenes. This is because most of the points are concentrated in the body torso, which is already the most stable tracked body part. We show snippets of the reconstructed body mesh for both scenes in Fig.~\ref{fig:humanMesh}. \Blue{As further highlighted in the supplementary video, unless the frame sequencing is properly estimated, the fitted body motion contains noticeable jitters and loops. Our proposed approach is free from such artifacts and produces pleasing motion reconstruction.}

\section{Discussion and Conclusion}
We present a framework for dynamic human reconstruction from uncalibrated and unsynchronized and moving rolling shutter cameras in the wild. The asynchronous video streams trivially invalidate the commonly used geometric triangulation constraint. The key to our approach is the use of physics-based motion prior to joint spatiotemporally calibrate the cameras and reconstruct the observed feature points, as well as the rolling shutter scanning speed. We devise an incremental reconstruction and alignment that strictly enforces the motion prior during the optimization and a divide and conquer algorithm that speeds up the first algorithm many folds without noticeable loss of accuracy. 

For better visual interpretation of the reconstruction, we fit a statistical human mesh model of the observed videos. This fitting is constrained by the same motion prior in addition to the detected semantic cues in the images. We validate the significant benefit of our spatiotemporal bundle adjustment to the estimated body shape and motion over the synchronized capture assumption for both synthetic and real video sequences.

\Blue{While we showcase the reconstruction of human with distinctive appearance such that simple color histogram is sufficient to identify the person, our algorithm can directly benefit from recent exciting progress in human appearance descriptor learning~\cite{vo2020self,zhou2019learning,zhang2019densely} and leverage such descriptor for multiple person tracking in more complex scenes.}

\section*{Acknowledgement}
This research is supported by NSF CNS-1446601, ONR N00014-14-1-0595, Heinz Endowments ``Platform Pittsburgh'', Metro 21 grants, and an Adobe Research Gift. Minh Vo was partly supported by the 2017 Qualcomm Innovation Fellowship.

{\small
\bibliographystyle{ieee}
\bibliography{egbib}
}

\boldstart{Minh Vo} is a Research Scientist at Facebook Reality Lab. He is interested in developing large-scale dynamic human scene understanding systems in order to create virtual environments that are perceptually indistinguishable from reality. He is the recipient of the Qualcomm Innovation Fellowship (2017). He received his Ph.D. from The Robotics Institute at Carnegie Mellon University.\\

\boldstart{Yaser Sheikh} is the Director of the Facebook Reality Lab in Pittsburgh and is an Associate Professor at the Robotics Institute, Carnegie Mellon University. His research is broadly focused on machine perception of social behavior, spanning computer vision, computer graphics, and machine learning. His works have won Popular Science’s “Best of What’s New” Award, the Honda Initiation Award (2010), Best Student Paper Award (CVPR 2018), Best Paper Awards (WACV 2012), SAP (2012), SCA (2010), ICCV THEMIS (2009), Best Demo Award (ECCV 2016), and placed first in the MSCOCO Keypoint Challenge (2016), the Hillman Fellowship (2004). His research has been covered in popular press including NY Times, The Verge, Popular Science, BBC, MSNBC, New Scientist, Slashdot, and WIRED. He obtained his Ph.D. from University of Central Florida.\\

\boldstart{Srinivasa G. Narasimhan} is a Professor of Robotics at Carnegie Mellon University. His group focuses on imaging, illumination and light transport to enable applications in vision, graphics, robotics, agriculture, and medical imaging. His works have won Best Paper Award (CVPR 2019), Best Demo Award (ICCP 2015), A9 Best Demo Award (CVPR 2015), Marr Prize Honorable Mention Award (2013), FORD URP Award (2013), Best Paper Runner up Prize (ACM I3D 2013), Best Paper Honorable Mention Award (ICCP 2012), Best Paper Award (IEEE PROCAMS 2009), the Okawa Research Grant (2009), the NSF CAREER Award (2007), Adobe Best Paper Award (ICCVW 2007) and Best Paper Honorable Mention Award (CVPR 2000). He obtained his Ph.D. from Columbia University.
\end{document}